\documentclass[sigconf]{acmart}
\usepackage{booktabs}
\usepackage{subcaption}
\usepackage{amsmath}
\usepackage[capitalize,noabbrev]{cleveref}
\usepackage{float}
\usepackage{url}
\usepackage{microtype} 
\usepackage{enumitem}
\graphicspath{{figures/}}

\copyrightyear{2026}
\acmYear{2026}
\setcopyright{cc}
\setcctype{by}
\acmConference[ICAIF '26]{7th ACM International Conference on AI in Finance}{November 14--17, 2026}{Milan, Italy}
\acmBooktitle{7th ACM International Conference on AI in Finance (ICAIF '26), November 14--17, 2026, Milan, Italy}
\acmDOI{10.1145/3858063.3858855}
\acmISBN{979-8-4007-3087-0/2026/11}

\begin{document}

\title[Rule Forest]{Rule Forest: Interpretable Ensembles of LLM-Generated Questions for Rare-Outcome Prediction}

\author{Ben Griffin}
\orcid{0009-0000-7310-6074}
\affiliation{\institution{University of Oxford}\city{Oxford}\country{United Kingdom}}
\affiliation{\institution{Vela Research}\city{San Francisco}\country{United States}}
\email{ben.griffin@keble.ox.ac.uk}

\author{Aaron Ontoyin Yin}
\orcid{0009-0002-9142-5890}
\affiliation{\institution{Vela Research}\city{San Francisco}\country{United States}}

\author{Diego Vidaurre}
\orcid{0000-0002-9650-2229}
\affiliation{\institution{Centre de Recerca Matem\`atica}\city{Barcelona}\country{Spain}}

\author{Ugur Koyluoglu}
\orcid{0000-0002-6286-351X}
\affiliation{\institution{Oliver Wyman}\city{New York}\country{United States}}

\author{Joseph Ternasky}
\orcid{0009-0004-0532-3180}
\affiliation{\institution{Vela Research}\city{San Francisco}\country{United States}}

\author{Fuat Alican}
\orcid{0009-0007-6185-4658}
\affiliation{\institution{Vela Research}\city{San Francisco}\country{United States}}

\author{Yigit Ihlamur}
\orcid{0009-0008-7091-5309}
\affiliation{\institution{Vela Research}\city{San Francisco}\country{United States}}

\begin{abstract}
Many high-stakes screening decisions in finance amount to predicting a rare outcome from unstructured text, such as whether a founder will successfully build a large company, or which crowdfunded loan will fail to attract funding. When positives are this rare, simple models may be preferable to more complex ones. For example, a model that learns a weight for each feature may have too little signal to estimate those weights reliably, and also inspecting the model means reading a list of fitted coefficients rather than a simple rule that can be checked by eye. We introduce Rule Forest, which uses a large language model to generate simple YES/NO questions, rather than using it to directly predict outcomes, and combines the strongest questions into an equal-weight ``green-flags'' scorecard. This scorecard is as accurate as an equivalent model that learns a weight per question, within noise, and its advantage appears largest when positives are rarest. The result is an interpretable list of rules that a domain expert can edit and extend. When predicting founder success,  the fraction of shortlisted founders who succeed increases from about 1 in 50 to 1 in 10, and to about 1 in 6 with a handful of expert-written questions. A second dataset which focuses on identifying failed loans, Kiva microloans, shows the same qualitative pattern. Across both tasks, the lesson is the same: a language model is most useful not as the predictor but as a generator of features a person can read, edit, and improve, and when positives are scarce that interpretability costs no measurable accuracy.
\end{abstract}



\maketitle

\section{Introduction}

In many high-stakes decision-making settings, a rare outcome is predicted from data consisting of unstructured text, where an incorrect decision may be extremely costly, and therefore the decision making process must remain transparent. Our primary application is startup founder investing in venture capital (VC), where the task is to predict, from a founder's background, whether their company will go on to reach an exceptional outcome. However, quantitative data is limited when evaluating founders, and much of the remaining data lives in messy profile text (e.g., education or prior job descriptions). This problem is particularly challenging because only 1.9\% of founders achieve \$500M+ outcomes~\cite{xiong2024gptree}, yet those rare successes can often return an entire fund. We additionally study a second dataset, Kiva microloans, where the task is to identify loans that will fail to reach their funding goal, which only happens 4.4\% of the time in this dataset.

For these types of tasks, recent work has begun applying machine learning to improve the decision making process~\cite{arroyo2019assessment, dellermann2021finding, maarouf2025fusedllm}, but these models are often not interpretable to the people who use them. Given that the inputs are often free-form text, classical pipelines often convert the text into features such as TF-IDF vectors or dense embeddings, neither of which an expert can read directly. Instead, models used in high-stakes decision making should be both accurate and explainable. Here, we therefore use a large language model (LLM) to turn the raw text directly into human-readable binary features.

We propose \emph{Rule Forest}, an interpretable ensemble framework in which an LLM generates many simple YES/NO questions in natural language. Each question acts as a weak heuristic applied to a founder's profile text, and combining the answers gives a transparent, precision-oriented predictor. The model reads as a ``green-flags'' scorecard, where scoring enough independent positive answers indicates a higher chance of success. This is related to classical ensemble ideas, but Rule Forest derives diversity from semantically distinct questions rather than from resampling or model variation \cite{dietterich2000ensemble,breiman1996bagging}.

Concretely, the LLM generates a pool of diverse YES/NO questions (e.g., ``Did the founder study computer science at a top-20 university?''); we retain the top $N$ and predict success when at least $T$ are answered \texttt{YES}. This unit-weight vote is closely related to classic \emph{unit-weighted} linear models~\cite{dawes1979robust}, and it works well precisely when positive examples are rare and coefficients for the selected questions are hard to estimate, such as in VC.

Our evaluation tests whether the LLM is more useful as a generator of interpretable features than as an end-to-end predictor, and separates Rule Forest's two design choices, the feature representation and the aggregation rule. We compare against direct prompting baselines (i.e., the LLM as a single-shot predictor). To isolate the aggregation rule, we fit a learned-weight classifier on Rule Forest's own binary question features, testing learned weights against the unit-weight vote. In order to isolate the feature representation, we run classical pipelines on alternative representations such as hand-engineered features and dense text embeddings.

We make the following contributions: $(1)$ Building on text features generated by LLMs~\cite{balek2025llm} and classical unit-weighted models~\cite{dawes1979robust,einhorn1975unit}, we introduce a method for rare-outcome prediction in finance, which turns unstructured text into a set of human-readable YES/NO questions, combined by an equal-weight vote into a scorecard. $(2)$ We show that this equal-weight vote is as accurate as a learned-weight classifier, within noise, when positive examples are scarce, with the same pattern appearing qualitatively on a second finance dataset. $(3)$ We demonstrate that, because every rule is human-readable, a domain expert can add their own questions to the scorecard and, in our experiment, raise its observed precision, with each added rule contributing to the score in a way the expert can see directly.

\begin{figure*}[t]
    \centering
    \includegraphics[width=1.0\textwidth]{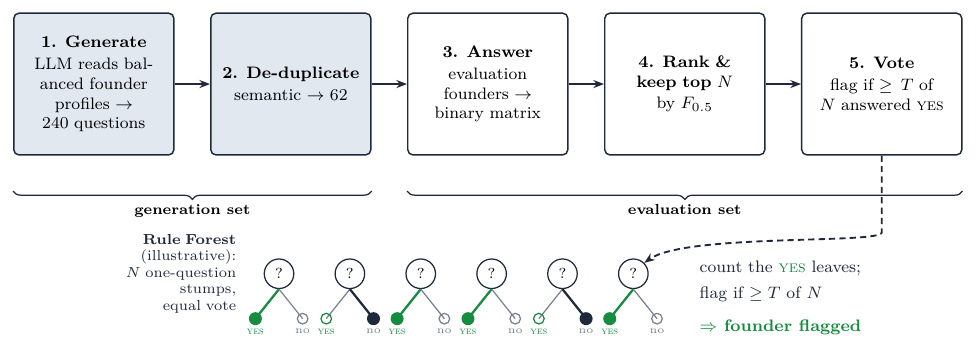}
    \Description{A pipeline of five boxes joined by arrows. The first two are shaded and grouped under "generation set": 1, Generate, where an LLM reads balanced founder profiles and writes 240 questions; and 2, De-duplicate, where semantic de-duplication leaves 62. The last three are grouped under "evaluation set": 3, Answer, where the questions are answered for the evaluation founders to form a binary matrix; 4, Rank and keep the top N by F0.5; and 5, Vote, where a founder is flagged if at least T of the N are answered YES. A dashed arrow from the Vote box points to an illustrative row of six one-question decision stumps, each a question mark with a YES leaf and a no leaf. For one founder, four stumps land on YES and two on no. The text beside them reads: count the YES leaves, flag if at least T of N, founder flagged.}
    \caption{Overview of the Rule Forest pipeline. An LLM is used to generate  YES/NO questions using a held-out question generation set of balanced founders, which can then be semantically de-duplicated. Each retained question is applied to the evaluation founders to form a binary question-response matrix; questions are ranked by $F_{0.5}$ and a founder is predicted successful when at least $T$ of these $N$ are answered \textsc{yes}. Equivalently, the model is a forest of one-question decision stumps combined by an equal-weight vote (right).}
    \label{fig:rrf_pipeline}
\end{figure*}

\section{Related Work} \label{sec:related_work}

\paragraph{Interpretable rule and tree frameworks.}
Our work sits within a line of research on converting opaque models into human-readable rules or trees, which predates LLMs. Earlier approaches extracted decision trees or IF--THEN rules from trained neural networks \cite{zilke2016deepred}. More recent work uses LLMs to propose rules directly. Founder-GPT \cite{xiong2023founder} combines LLM self-play with multi-expert prompting to assess founder--idea fit, but returns narrative judgements rather than a compact rule set. GPTree \cite{xiong2024gptree} instead uses a tree-structured decomposition of the input space. We share the goal of transparent reasoning, but instead of a tree or a written rationale we use a flat set of independent YES/NO questions combined by a vote. Balek et al.~\cite{balek2025llm} also use an LLM to extract interpretable text features in place of embeddings, but combine them with learned classifiers; we instead combine them with an equal-weight vote, following the classical case for unit weights~\cite{dawes1979robust,einhorn1975unit}, which is more suited to domains with extremely rare positive cases.

\paragraph{Startup-success prediction and black-box LLM ensembles.}
Traditionally, quantitative approaches to startup-success prediction used manually engineered features with conventional classifiers such as boosted trees and support vector machines \cite{arroyo2019assessment}; other work adds a human-in-the-loop, fusing crowd judgements with machine predictions \cite{dellermann2021finding}. More recent black-box approaches combine structured startup features with text embeddings \cite{maarouf2025fusedllm}, or aggregate the final forecasts of multiple LLMs (e.g., by taking the median), without combining the models' intermediate reasoning \cite{schoenegger2024siliconcrowd}.

The closest existing founder-success benchmark, VCBench~\cite{
chen2026vcbench}, evaluates predictors at a $9\%$ success rate; we instead target the harder, more realistic
regime in which fund-returning outcomes are far rarer ($1.9\%$)~\cite{xiong2024gptree}, which is precisely where the choice of combination rule begins to matter.

\paragraph{Credit and lending prediction.} Beyond founder evaluation, LLMs have been applied to other finance tasks such as generating credit-risk reports~\cite{teixeira2023credit} and scoring credit risk directly from loan descriptions~\cite{sanzguerrero2025credit}. Interpretability is extremely important in these domains, with recent work examining LLM-based credit classifiers with post-hoc feature-attribution methods~\cite{almarri2025interpreting}; our second dataset, Kiva microloans, is drawn from this domain.

\section{Datasets}

\subsection{Founder Success Dataset}

We constructed a dataset of 9,192 founder profiles from a combination of licensed company data and professional-background data obtained via third-party providers. Each profile corresponds to a U.S.-based company founded between 2010 and 2016. We restrict to U.S. startups to ensure consistent success definitions and comparable founder signals.

To support open research while preserving commercial value, we transformed the raw historical data into an anonymised, NLP-style format using a proprietary LLM-powered pipeline. Each summary described a founder’s background using structured natural language derived from metadata such as education, work history, and prior startup experience, based solely on information available at the time investment decisions were made. This transformation ensures that the released dataset protects proprietary information while preserving core predictive content. Our anonymised dataset, code, and full prompt templates are available at \url{https://github.com/Vela-Research/rule-forest-icaif}.

The dataset includes companies that had raised between \$100K and \$4M as of the investment-decision point, typical of an early-stage pool that a seed-stage investor might look to invest in. Following VCBench~\cite{chen2026vcbench}, a founder is labelled \texttt{successful} if the company was acquired or had an IPO above \$500M valuation, or raised more than \$500M in funding, and \texttt{unsuccessful} if the company raised \$100K--\$4M at inception but did not achieve an exit, IPO, or substantial follow-on funding within eight years of founding.

\subsection{Kiva Microloan Dataset}

Our second dataset is from Kiva, an online microfinance platform on which lenders crowdfund loans. From a public Kiva data snapshot~\cite{kiva_snapshot} of roughly 1.4 million loans (posted 2006--2018), we take a uniform random sample of 5{,}000 loans that reached a funding decision and carry an English-language narrative. The task is to predict whether a loan reaches the end of its fundraising window without being fully funded. We label a loan positive if it expired unfunded and negative if it was fully funded. Funding failure is rare (4.4\%), mirroring the low-base-rate, hard-screening regime of the founder task, but in the credit and financial-inclusion domain.

\section{Methods}

Rule Forest converts unstructured text into a small set of human-readable YES/NO questions and combines them with a simple vote. The pipeline has three stages. $(1)$ An LLM generates candidate questions from a set of labelled examples; $(2)$ each question is applied to every example to form a binary matrix; and $(3)$ the top-ranked questions are combined by a unit-weight vote.

Since question generation uses labelled examples, it is a form of supervised feature construction, so we keep the examples used to generate questions separate from those used for training and evaluation. Just under 1,000 founders are held out from evaluation entirely, 497 used exclusively for question generation and 491 as a validation set used only to screen out weak questions before scoring. Both are deliberately enriched with successful founders, since generating and screening questions needs enough positive examples to be informative. The remaining 8,204 founders form the evaluation set, at the 1.9\% base rate that matches the real-world success rate, and are used for model selection and performance estimation by cross-validation. Question generation is run once on the held-out set and then fixed, which isolates the pipeline from the randomness of the generation step and avoids regenerating questions inside every cross-validation fold, a step that would be prohibitively expensive. Figure~\ref{fig:rrf_pipeline} summarises the pipeline described in full detail in the following sections.

\subsection{LLM-Guided Question Generation}
\label{section:question_generation}

For the founder dataset, each prompt shows the LLM 20 founders (10 successful and 10 unsuccessful where possible) and asks for 10 YES/NO questions, at temperature 0. This is repeated 25 times, producing  240 usable candidate questions.

At this stage, an optional preprocessing step filters out questions that are semantically very similar, measured by cosine similarity between \texttt{all-MiniLM-L6-v2} sentence embeddings, reducing API calls to the LLM to answer the questions without affecting the final model. A later step (\Cref{section:question_refinement}) removes questions that are predictively similar, which would also capture those that are semantically similar. This filtering also shows that the underlying set of distinct heuristics is small. We found that 240 generated questions reduced to only 62 after semantic filtering. 
The retained questions are behaviourally diverse rather than paraphrases, with their founder-response vectors differing on about 40\% of founders on average. \Cref{fig:genprompt} shows the generation prompt and \Cref{tab:example_questions} shows a sample of the resulting questions, with two illustrative rules an expert added.

\begin{table}[ht]
\centering
\caption{A sample of Rule Forest founder evaluation questions. Each question is a plain YES/NO criterion, so a reader can compare any answer with the founder's profile. The first 8 questions are LLM-generated, and the bottom two are illustrative rules generated by a domain expert.}
\label{tab:example_questions}
\small
\begin{tabular}{p{0.74\columnwidth}c}
\toprule
\textbf{Question} & \textbf{Source} \\
\midrule
Does the founder hold a PhD from a prestigious institution? & LLM \\
Does the founder have experience in venture capital? & LLM \\
Does the founder have advanced degrees in a technical field? & LLM \\
Has the founder served on an advisory board for a large company? & LLM \\
Has the founder held a C-level position in a previous startup? & LLM \\
Does the founder have experience in financial services? & LLM \\
Is the founder based in California? & LLM \\
Does the founder have a strong personal brand from media coverage? & LLM \\
\midrule
Has the founder sold a prior startup for over \$50M? & Expert \\
Has the founder's prior startup been acquired by a reputable public company? & Expert \\
\bottomrule
\end{tabular}
\end{table}

\begin{figure}[ht]
\centering
\fbox{\parbox{0.93\columnwidth}{\small\textbf{System.} You are a VC analyst evaluating founders. Generate objective YES/NO questions from founder summaries, covering education, work history, leadership, and relevant experience.\par\smallskip \textbf{User.} Generate 10 YES/NO questions answerable from the anonymised founder summaries below. Each must be objective and specific (education, roles, founding experience), one per line, with no numbering or explanation.\par\smallskip \textit{[20 anonymised founder summaries]}}}
\Description{A boxed, condensed copy of the question-generation prompt. A system message casts the LLM as a VC analyst writing objective YES/NO questions about founders, and a user message asks for 10 such questions from 20 anonymised founder summaries, one per line.}
\caption{The question-generation prompt (condensed; the full template is in the released code).}
\label{fig:genprompt}
\end{figure}

\subsection{Question Evaluation, Redundancy Removal, and Ranking}\label{section:question_refinement}

After using the LLM to generate candidate questions, we evaluate them by asking an LLM to answer each question for each founder with a YES/NO response, generating a binary feature vector per question. The prompt allows only a YES or NO answer, with no option for missing information, and any reply other than YES, including one that cannot be read (0.1\% of answers in the main analysis), counts as NO. To limit cost, candidate questions are first answered on the validation founders, and those with very low precision are discarded before the survivors are scored on the full evaluation set. This is a cost saving choice rather than a modelling one, since with an unlimited budget every question could be scored directly. Since the validation set is enriched with successes, each question's precision is estimated from enough positive examples to be a stable reading, and the screen is loose enough that only questions performing no better than chance are removed. In our run it pruned 12 questions, substantially reducing the number of API calls needed.

For the vote ensemble, we encourage diversity by removing questions with similar predictions. Specifically, for two binary response vectors $\mathbf{a}, \mathbf{b} \in \{0,1\}^n$, we compute the Hamming distance
\begin{align*}
d_H(\mathbf{a}, \mathbf{b}) = \frac{1}{n} \sum_{i=1}^{n} \mathbf{1}[\mathbf{a}_i \ne \mathbf{b}_i]
\end{align*}
between them.

If the distance falls below a similarity threshold (here we use $0.15$), we retain only the question with the higher precision. The remaining questions are ranked by their individual $F_{0.5}$ score. Redundancy removal and ranking are recomputed inside each training fold, so no question's inclusion or position in the ranked list depends on test-fold labels. This ranked list of questions defines the candidate pool which is used to construct the ensemble.

\paragraph{Computational cost.} Answering the retained questions for every founder is the major cost of the pipeline, requiring one LLM call for each founder and for each question. On the full evaluation set this takes roughly 67 hours sequentially, at approximately US\$21 in API calls (GPT-4o-mini, \$0.15 and \$0.60 per million input and output tokens), compared with 1--2 hours for the direct prompting baselines. The step parallelises naturally across founders and questions, but Rule Forest is clearly the more expensive option.

\subsection{Ensemble Construction and Tuning}
\label{section:ensemble_prediction}

Our main predictor is a unit-weight vote ensemble with two hyperparameters, the number of retained questions $N$ and the voting threshold $T$. For example $i$, let $x_{ij}\in\{0,1\}$ denote the response to question $j$. The ensemble score over the top-$N$ ranked questions is
\[
s_i = \sum_{j=1}^{N} x_{ij},
\]
and we predict success when $s_i \ge T$.

We tune $(N,T)$ in the inner loop of nested cross-validation to optimise the target objective. Increasing $N$ expands the rule set available to the ensemble, while $T$ controls the precision--recall trade-off. Giving every retained question an equal vote keeps the model simple enough for an expert to read, edit, and use to inform high-stakes decisions involved in venture capital.

\subsection{Learned-weight and classical baselines}
\label{sec:learned_weights}

Using the full binary feature matrix developed above, we compare Rule Forest against an elastic-net logistic regression over the same questions, to examine whether we are trading accuracy for interpretability by forcing uniform weights. The elastic net does its own selection through the penalty on $\mathbf{w}$, so it is given all the questions rather than the vote's top-ranked few; \Cref{fig:mechdetail} isolates the aggregation rule directly by giving both combiners an identical question set.

After question generation and evaluation, each example $i$ is represented by a binary response vector $\mathbf{x}_i \in \{0,1\}^{Q}$, where $x_{ij}=1$ indicates that question $j$ is answered \emph{YES} for example $i$. We fit
\[
p(y_i=1\mid \mathbf{x}_i)=\sigma(\mathbf{w}^{\top}\mathbf{x}_i + b),
\]
using an elastic net penalty on $\mathbf{w}$, where its hyperparameters are tuned via nested cross-validation, matching the protocol of Rule Forest.

\paragraph{Classical baselines.}

We also compare against two traditional feature representations, hand-engineered features~\cite{arroyo2019assessment} and dense text embeddings~\cite{maarouf2025fusedllm}, which isolates the value of the LLM-generated question representation. The \emph{Engineered ML} approach extracts a fixed set of deterministic features from each summary using keyword and pattern rules, followed by an elastic-net logistic regression. The \emph{Embeddings} approach encodes each summary with a frozen all-MiniLM-L6-v2 sentence encoder and classifies the resulting 384-dimensional vector with XGBoost.

\paragraph{Prompting and other LLM-based baselines.} Finally, we also compare Rule Forest against other LLM-based approaches, including using the LLM directly as a predictor. In zero-shot prompting, the LLM labels a founder as successful or unsuccessful from the profile alone. In few-shot prompting, it sees 4 labelled examples (two successful and two unsuccessful) drawn from profiles outside of the evaluation set. We further investigate two recent approaches that use LLMs to generate features used for prediction. GPTree~\cite{xiong2024gptree} generates an LLM-built decision tree, and Reasoned Rule Mining (RRM)~\cite{preuveneers2025reasoned} mines IF--THEN rules from an LLM's reasoning traces and combines the resulting scores by a learned weighted sum on the log-odds scale.

\subsection{Evaluation Protocol}
\label{sec:eval_protocol}
Unless stated otherwise, all results are computed on the evaluation set using 10-fold nested cross-validation repeated 10 times with different random splits (100 outer test folds total). In each outer fold, models are fit on the training split and evaluated on the held-out test fold; hyperparameters and, where applicable, decision thresholds are selected using only the inner folds. Prompting-based baselines are not trained and are therefore evaluated once on the evaluation set, without cross-validation or inner-loop tuning.
We report precision, recall, and $F_{0.5}$, the weighted harmonic mean $F_\beta = (1+\beta^2)\,PR/(\beta^2 P + R)$ at $\beta = 0.5$, which weights precision more heavily than recall. False positives are costly in venture capital, since each shortlisted founder consumes partner time, but a shortlist with too few candidates is no use either.

The Kiva experiments, which use the same $10\times10$ nested cross-validation, explore whether, in a second domain, the equal-weight vote is again as accurate as a learned-weight model, within noise, when positives are scarce. Questions are generated by Gemini 2.5 Flash on the training data, and the equal-weight vote and elastic-net combiners on those questions are the same as in the founder setup. All prompts, the precomputed question--response matrices, the cross-validation and bootstrap seeds, and the search grids used by the released producers are available at \url{https://github.com/Vela-Research/rule-forest-icaif}.

\begin{table*}[t]
\centering
\caption{Founder evaluation-set performance. Trained models and the LLM tree/rule-mining baselines report mean~$\pm$~SD over $10\times10$ nested cross-validation; prompting baselines are evaluated once. GPTree and RRM are re-evaluated on our dataset under this protocol, with decision thresholds selected in-fold.}
\label{tab:unweighted-perf}
\begin{tabular}{llccc}
\toprule
\textbf{Model} & \textbf{Approach} & \textbf{Precision} & \textbf{Recall} & $F_{0.5}$ \\
\midrule
Rule Forest              & LLM + unit vote      & 0.103 $\pm$ 0.015 & 0.080 $\pm$ 0.013 & 0.097 $\pm$ 0.014 \\
Engineered ML    & Engineered features  & 0.071 $\pm$ 0.013 & 0.055 $\pm$ 0.020 & 0.066 $\pm$ 0.014 \\
LogReg (EN)      & Learned weights      & 0.112 $\pm$ 0.022 & 0.066 $\pm$ 0.011 & 0.098 $\pm$ 0.017 \\
Embeddings       & Embeddings + XGBoost & 0.121 $\pm$ 0.059 & 0.014 $\pm$ 0.008 & 0.047 $\pm$ 0.025 \\
\midrule
o3               & Few Shot             & 0.077 & 0.196 & 0.088 \\
GPT-4o-mini      & Zero Shot            & 0.079 & 0.139 & 0.087 \\
\midrule
GPTree      & LLM decision tree & 0.068 $\pm$ 0.019 & 0.095 $\pm$ 0.036 & 0.071 $\pm$ 0.021 \\
RRM         & LLM rule mining   & 0.080 $\pm$ 0.014 & 0.102 $\pm$ 0.025 & 0.082 $\pm$ 0.012 \\
\bottomrule
\end{tabular}
\end{table*}

\section{Experiments}

We now report results on both datasets. The founder dataset is our main evaluation and carries the baseline comparisons and the expert-in-the-loop study; the Kiva dataset explores whether our main finding, that the equal-weight vote holds up as positives grow scarce, carries over to a second, credit-domain task.

\subsection{Experimental Results} \label{sec:exp_res}

\Cref{tab:unweighted-perf} shows accuracy scores across models for founder success prediction. Rule Forest ($0.097$) and the learned-weight approach on the same binary feature matrix ($0.098$) achieve the highest $F_{0.5}$, ahead of direct LLM predictions (best $0.088$, o3 few-shot), LLM-native approaches (best $0.082$, RRM), and classical baselines (best $0.066$, Engineered ML), though embeddings reach higher precision at much lower recall. Although these scores are low in absolute terms, this is not unexpected given how rare positives are (under $2\%$) and how hard founder success is to predict.

\begin{figure}[t]
  \centering
  \includegraphics[width=\columnwidth]{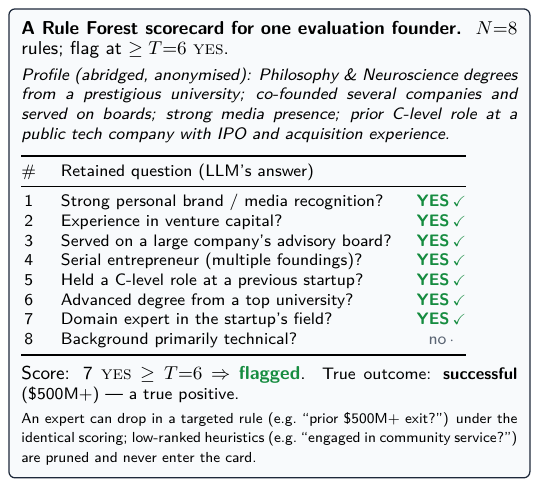}
  \Description{A scorecard card for one evaluation founder, with N equal to 8 rules and a flag at T equal to 6 YES answers. A short anonymised profile mentions degrees from a prestigious university, several companies co-founded, board roles, a strong media presence and a prior C-level role at a public tech company. A table lists eight questions with the LLM's answer to each: strong personal brand, venture capital experience, a large company's advisory board, serial entrepreneur, a C-level role at a previous startup, an advanced degree from a top university and domain expertise are YES, and a primarily technical background is no. The score line reads 7 YES, at least 6, so the founder is flagged; the true outcome is successful, a true positive. A note says an expert can add a targeted rule, and low-ranked questions are pruned and never enter the card.}
  \caption{A Rule Forest prediction is a readable ``green-flags'' scorecard; the founder is predicted successful when at least $T$ of the $N$ retained YES/NO questions are answered YES. The example is a real evaluation founder and the LLM's answers, with the profile and question wording shortened.}
  \label{fig:scorecard}
\end{figure}

The learned baselines separate Rule Forest's two design choices. A founder-level bootstrap (10,000 resamples) cannot separate Rule Forest from the elastic net ($\Delta = -0.001$, $95\%$ confidence interval $[-0.032, +0.035]$) or from o3 few-shot, the strongest direct LLM prediction ($\Delta = +0.010$, $[-0.034, +0.055]$). Rule Forest's primary advantage is instead its simplicity. \Cref{fig:scorecard} shows an example of the final scorecard it produces, which is a readable, editable list of rules an expert can improve directly.

We also note that differences this small are hard to establish statistically at a 1.9\% base rate, but the returns to a slightly better shortlist in venture capital are large. Of the 8,204 evaluation founders, $158$ succeed. Rule Forest's shortlist of about $120$ founders contains roughly $13$ of them; to find those same successes by screening at random, a partner would review about $660$ profiles, so one shortlisted founder in ten succeeds rather than one in fifty. Since a single $\$500$M+ outcome can return an early-stage fund, what matters is how many such founders a partner finds for a given number of profiles they review.

\subsection{Impact of Expert-in-the-Loop Refinement}
\label{sec:expert}

We also consider an expert-in-the-loop addition, where 8 expert-written questions replace 8 of the LLM questions, so the list keeps its size, and we treat them like LLM-generated ones (evaluated, scored, filtered, and ranked by the same procedure), except that the similarity threshold is $0.01$ rather than $0.15$. Three expert questions are answered YES for only 1--4\% of founders. At $0.15$, the Hamming filter would count them as near-duplicates of some of the strongest LLM questions, which are also mostly answered NO, and remove those LLM questions. The eight expert questions, two of which are shown in \Cref{tab:example_questions}, were written by a general partner at an early-stage venture firm who has invested in dozens of startups, and they encode the signals that an investor looks for when evaluating founders.

By adding these expert questions, both the precision of the augmented Rule Forest increased ($0.103\pm0.015$ to $0.164\pm0.020$) and the $F_{0.5}$ ($0.097\pm0.014$ to $0.125\pm0.015$), while the recall decreased ($0.080\pm0.013$ to $0.064\pm0.009$). Using the same bootstrapping approach as in \Cref{sec:exp_res}, the improvement over Rule Forest alone is not statistically separable (precision $\Delta=+0.061$, CI $[-0.007, +0.135]$). Against o3 few-shot, which flags 402 founders without ranking them, we compare precision at equal list sizes. At about 60 founders, the scorecard's own list size, its precision is $0.150$ against $0.077$ ($\Delta=+0.073$, CI $[+0.003, +0.153]$), while at 402 founders the two are equal. In other words, the expert questions are more targeted, pushing the ensemble towards higher precision. Given the same 58 questions, the elastic net gains recall but not precision ($0.103$ against $0.164$, $\Delta=+0.061$, CI $[-0.009, +0.139]$). Since the expert questions replace LLM ones, the gain does not come from adding questions, although the lower similarity threshold leaves the expert arm a larger pool to choose from ($47$--$49$ questions against $21$--$24$). This shows the advantage of having a simple, interpretable model, where a domain expert can directly add rules to the final Rule Forest scorecard themselves, quickly inspecting, adding, replacing, or removing rules from a list of a few dozen. Overall, the expert-in-the-loop extension shows an observed relative gain of roughly 60\% in precision from eight human-written rules, while keeping recall high enough to leave a usable shortlist, representing a cheap but promising form of human--AI collaboration.

\subsection{Robustness and ablations}

\paragraph{Robustness} \Cref{tab:ablations} shows how Rule Forest performs when evaluated under alternative preprocessing choices. Before scoring, we remove near-duplicate questions with a Hamming-similarity filter (threshold 0.15); the similarity metric has little effect on the $F_{0.5}$ score, but raising the threshold to 0.20 removes too many questions and lowers the $F_{0.5}$ score to $0.059$. Hamming is our default because it was fixed before evaluation.

Hyperparameter selection is consistent. Across the 100 outer folds the inner loop selects a median of 8 questions with a voting threshold of 6, and never reaches the edge of the search grid. The selected $N$ and $T$ vary from fold to fold, but the fraction of questions required to fire stays close to three quarters (median $T/N=0.75$, interquartile range $0.71$--$0.80$), so repeated tuning converges on the same operating point rather than on unrelated scorecards.

\paragraph{Question generator.} We found that the LLM used to generate questions had the largest impact in terms of overall $F_{0.5}$ score. While GPT-4o-mini is the best model with $F_{0.5}=0.097$, alternative models scored $0.079$ for Claude Sonnet 4.5 and $0.048$ for GPT-5.2. For GPT-5.2, this is a $103\%$ relative gap that would likely be economically meaningful for venture capital decision-making, and a founder-level bootstrap of the gap gives $\Delta = +0.050$ with a $95\%$ confidence interval of $[+0.005, +0.097]$. Although GPT-5.2 is the more powerful reasoning model, GPT-4o-mini scores higher, which tracks question diversity, given that after deduplication, GPT-5.2 produced far fewer distinct questions. Regardless, we found that whichever model generates the questions, the equal-weight vote matches a learned-weight elastic net over the same questions, within noise (\Cref{tab:ablations}), so the editable scorecard costs no measurable accuracy, and \Cref{sec:count_effect} shows the relative performance of the equal-weight vote improving as positives grow scarce.

\begin{table}[t]
\centering
\caption{Ablation of Rule Forest pipeline components on the founder evaluation set ($F_{0.5}$; one component varied at a time, others fixed; default in bold). For each generator we also report an elastic net over the same binary questions.}
\label{tab:ablations}
\begin{tabular}{llc}
\toprule
\textbf{Component} & \textbf{Setting} & \textbf{$F_{0.5}$} \\
\midrule
Similarity metric & \textbf{Hamming} & \textbf{0.097} \\
                  & Jaccard & 0.093 \\
                  & Cosine clustering & 0.094 \\
\midrule
Hamming threshold & 0.10 & 0.088 \\
                  & \textbf{0.15} & \textbf{0.097} \\
                  & 0.20 & 0.059 \\
\midrule
Question generator & \textbf{GPT-4o-mini (vote)} & \textbf{0.097} \\
                   & GPT-4o-mini (elastic net) & 0.098 \\
                   & Claude Sonnet 4.5 (vote) & 0.079 \\
                   & Claude Sonnet 4.5 (elastic net) & 0.066 \\
                   & GPT-5.2 (vote) & 0.048 \\
                   & GPT-5.2 (elastic net) & 0.050 \\
\bottomrule
\end{tabular}
\end{table}

\subsection{When counting beats weighting}
\label{sec:count_effect}

Rule Forest combines its questions with a simple equal-weight vote rather than learning a weight for each one. To see when a learned-weight elastic net does not beat the vote on the founders dataset (\Cref{tab:ablations}), we fit both combiners on the evaluation founders under paired nested cross-validation while varying the number of successes in the sample. Since successes equal sample size times base rate, each sweep holds either the base rate, the sample size or the number of successes fixed.

\begin{figure*}[t]
  \centering
  \includegraphics[width=\linewidth]{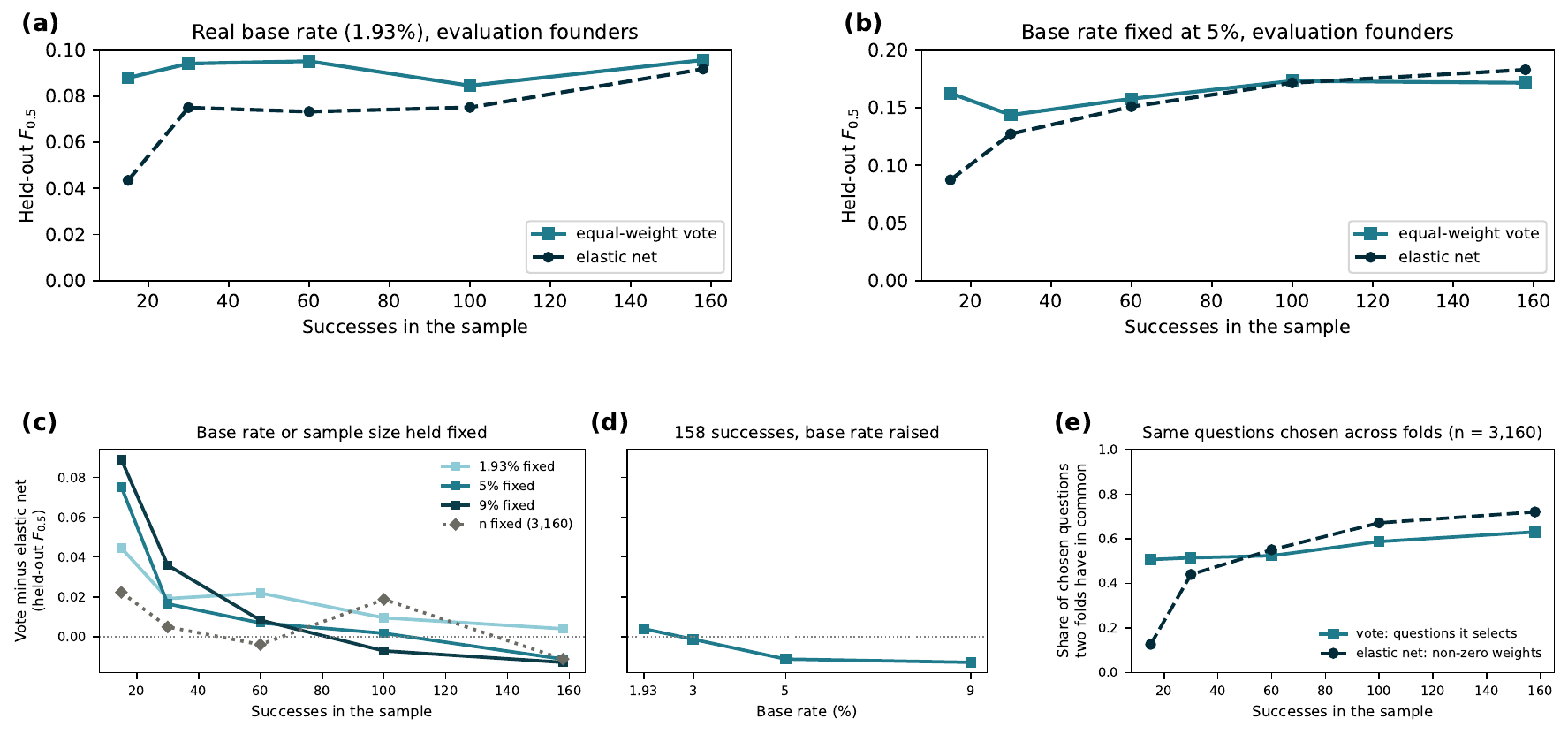}
  \Description{Five line charts. Panel a, real base rate of 1.93 percent: held-out F0.5 against successes in the sample from 15 to 158; the equal-weight vote stays near 0.09 while the elastic net rises from about 0.04 to about 0.09, staying below the vote. Panel b, base rate fixed at 5 percent: the vote starts well above the elastic net at 15 successes, the two lines meet near 100 successes, and the elastic net is slightly higher at 158. Panel c: vote minus elastic net against successes, with four lines for base rates of 1.93, 5 and 9 percent and for a fixed sample of 3,160 founders; every line is highest at 15 successes and falls to around zero by 158. Panel d: with all 158 successes kept, vote minus elastic net against base rates of 1.93, 3, 5 and 9 percent, falling from slightly above zero to slightly below. Panel e, sample of 3,160 founders: the share of chosen questions two folds have in common rises from about 0.5 to 0.6 for the vote and from about 0.1 to 0.7 for the elastic net, which is lower than the vote with few successes and higher from about 60 successes.}
  \caption{Held-out $F_{0.5}$ of the equal-weight vote and the elastic net as the number of successes in the sample varies, under paired nested cross-validation on the evaluation founders. (a) At the real 1.9\% rate the vote leads at every count, up to the 158 successes of our evaluation set. (b) With the rate fixed at 5\%, the elastic net catches up at about 100 successes. (c)  Vote minus elastic net with the base rate or the sample size held fixed. (d) Vote minus elastic net with all 158 successes kept and the base rate raised by removing unsuccessful founders. (e) Share of chosen questions two folds have in common. For the vote these are its selected questions, and for the elastic net the questions with non-zero weight. Each point is the mean of 10 subsamples, except the full-data point.}
  \label{fig:mechdetail}
\end{figure*}

\Cref{fig:mechdetail}a uses the true 1.9\% rate of successful founders, where the data includes 158 successes. \Cref{fig:mechdetail}b fixes the rate at 5\% by subsampling the evaluation founders, so no founder used for question generation is included.

At the 1.9\% base rate (\Cref{fig:mechdetail}a), the vote stays roughly flat, from $F_{0.5}=0.088$ to $0.096$, while the elastic net rises from $0.043$ to $0.092$, so the vote leads at every count and its margin is largest when successes are scarcest. This matches the evaluation set results (\Cref{tab:unweighted-perf}). The base rate we choose for the evaluation set, 1.9\%, is selected because it matches the real-world success rate~\cite{xiong2024gptree}. However, if we allow a higher base rate, then as the number of successful founders increases, the elastic net catches up with the vote and slightly overtakes it (\Cref{fig:mechdetail}b). The vote's lead is largest at 15 successes whether the base rate or the sample size is held fixed (\Cref{fig:mechdetail}c), whereas with all 158 successes it turns into a small deficit as the base rate rises (\Cref{fig:mechdetail}d). The vote's lead therefore comes mainly from having few successes. With 158 successes, the elastic net slightly overtakes it at enriched rates such as the 9\% used by VCBench~\cite{chen2026vcbench}, but at the real 1.9\% rate the vote never falls behind.

When successes are few, which questions the elastic net gives a non-zero weight also changes from fold to fold. With 15 successes, two of the elastic net's folds have on average 13\% of these questions in common, while two of the vote's folds have 51\% of their selected questions in common (\Cref{fig:mechdetail}e). While these results are consistent with the classical case for unit weights~\cite{dawes1979robust,einhorn1975unit}, they do not isolate a single cause since both combiners overfit when successes are few. However, the results suggest that the elastic net overfits more, with a larger gap between training and held-out $F_{0.5}$ for 18 of the 20 combinations of sample size and base rate in \Cref{fig:mechdetail}.

\subsection{\texorpdfstring{Generalisation to Kiva microloans}{Generalisation to Kiva microloans}}
\label{sec:kiva_generalisation}

\begin{figure}[t]
  \centering
  \includegraphics[width=\columnwidth]{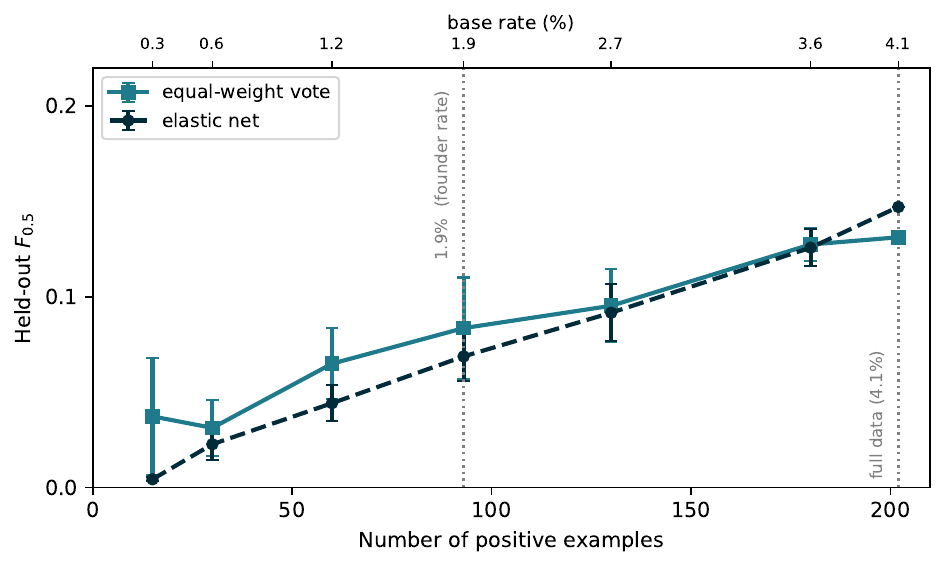}
  \Description{A line chart of held-out F0.5 against the number of positive Kiva loans, from 15 to 202, with the matching base rate from 0.3 to 4.1 percent on a top axis and 95 percent interval bars on each point. Dotted vertical lines mark the founder rate of 1.9 percent and the full data at 4.1 percent. The equal-weight vote is above the elastic net at every count below the full data, most clearly at 15 positives, where the elastic net is near zero. The two lines are close at 3.6 percent, and the elastic net is higher at the full data.}
  \caption{Comparison $F_{0.5}$ score of equal-weight vote and elastic net approaches on Kiva microloans as the number of positive examples is varied, with the resulting base rate on the top axis. The elastic net leads at the full-data base rate, but the equal-weight vote approach leads at every lower rate, including the founder base rate $1.9\%$ (dotted line). Points are $10\times10$ nested cross-validation over ten subsample draws ($95\%$ CIs); the full-data point is a single draw.}
  \label{fig:mechanism}
\end{figure}

This pattern is not specific to founders. On Kiva microloans, where the task is to identify loans that will expire without reaching their funding goal, we keep every negative and vary only the number of positive examples, so the base rate falls as positives are removed. At Kiva's full-data base rate the learned-weight elastic net slightly outperforms the vote, but as positives become scarcer the vote overtakes, leading at every lower rate we test, including our $1.9\%$ founder rate (\Cref{fig:mechanism}). The gaps are small, so we present this as a qualitative replication rather than a confirmatory test. What it shows is that the pattern is not an artefact of the founder data, since the equal-weight vote again holds its own as positives grow scarce in a different domain and on different text.

\section{Discussion} \label{sec:discussion}

Practitioners applying machine learning to rare-outcome prediction often face a familiar tradeoff, where a more accurate model tends to be a more complex one. On both of our datasets, this tradeoff largely disappears once positive examples are scarce. Rule Forest turns messy founder text into a set of YES/NO questions and combines them by an equal-weight vote, and we find that simply counting how many questions fire is as accurate as learning what each is worth, within noise, while leaving a scorecard an expert can read, edit, and improve by hand.

This is a classical argument for unit weights~\cite{dawes1979robust,einhorn1975unit}, and what we add is a way of applying it to unstructured text. The same pattern appears on Kiva, where the vote overtakes the elastic net as the positive rate approaches that of the founder task. The cost is that every retained question counts the same, so the vote cannot discount one that turns out to be weaker than the rest. Therefore, we recommend that in low base rate screening problems with only a few dozen training successes, an equal-weight vote is a strong default, and learning the weights begins to help only as the number of positives increases.

Rule Forest's real advantage over alternative approaches is transparency rather than accuracy. Rather than looking for a single prototypical founder, the ensemble asks whether enough green flags are present at all, so a technical founder with deep domain experience and a repeat operator with a strong network can both clear the same bar without sharing a single attribute. The ensemble is intentionally shallow, a handful of single-step rules combined by one global threshold, leaving a clear foundation for extensions. This division of labour suits text-heavy domains, where the LLM handles the hard translation from messy language into candidate features, while the final predictor stays a count of named rules a domain expert can inspect and extend. Founder screening has every constraint this design responds to. The signal lives in unstructured profile text, fund-returning outcomes can be too rare for per-question weights to be estimated reliably, and a partner will not act on a shortlist they cannot interrogate or amend.

\subsection{Limitations and Future Work} \label{sec:limitations}

Our approach has clear limitations. The question-generation step is stochastic and itself a black box, so reruns with a different prompt or temperature may generate different YES/NO questions.  The question answers are the LLM's reading of each profile and are not independently verified. Rule Forest's interpretability therefore lies in its readable aggregation, not in verified answers. The questions and the success labels are both derived from historical data, so they can encode selection or societal biases such as pedigree signals; keeping the rules human-readable is precisely what lets an expert inspect and remove such criteria before deployment. Moreover, any model fit to a specific population (e.g., US-based founders) may not perform as well in other regions or timeframes without re-validation. Finally, our results measure screening quality rather than investment performance, since real decisions also depend on access to the round, allocation, and price.

Future work could systematically study how prompt and temperature choices affect the generated questions, as well as biases inherent in the dataset or the LLM. A human audit could measure how often the LLM's answers match the profiles. Some attributes, such as how risky the target market is, are naturally graded rather than YES/NO, as in credit analysis, and could be asked as rating questions whose answers are cut at a threshold chosen on the training data. Approaches from anomaly detection~\cite{chandola2009anomaly} and imbalanced learning~\cite{he2009imbalanced}, such as the adversarial models used for credit-card fraud detection~\cite{fiore2019using}, could be used to select which questions enter the scorecard. Furthermore, Rule Forest could also be extended with richer rule structures, such as short AND/OR logic or shallow trees, that capture interactions a flat scorecard cannot. However, each extension risks sacrificing some of the simplicity and readability that motivate the method for potential accuracy gains, and may be better suited to different stages of the investment pipeline with higher base rates.

\section{Conclusion}

We presented Rule Forest, a method that turns unstructured text into a small set of human-readable YES/NO questions combined by an equal-weight vote. When positives are scarce, counting how many fire is as accurate as learning a weight for each, within noise. Therefore in rare-outcome screening such as founder evaluation, the LLM is most valuable as a generator of features a person can inspect, edit, and improve, rather than as the predictor itself.

\section{Ethical Considerations}

Founder data may contain sensitive information, so we release only anonymised summary text and report aggregate results. The learned rules can encode pedigree signals and may act as proxies for protected characteristics; whether a given signal is appropriate depends on the deploying institution, and we position the method as decision support under human oversight rather than automated decision-making.

\bibliographystyle{ACM-Reference-Format}
\bibliography{references}

\end{document}